%% file: dna1.0.tex
\documentclass{article}
\usepackage{dna1.0}

\PassOptionsToPackage{table}{xcolor}
\usepackage{etoolbox}
\usepackage[utf8]{inputenc}

\newtoggle{arxiv}
\toggletrue{arxiv}

\usepackage{booktabs}
\usepackage{graphicx}
\usepackage{enumitem}
\usepackage{wrapfig}
\usepackage{algorithm}
\usepackage{algpseudocode}

\usepackage{microtype}
\usepackage{amsmath}
\usepackage{colortbl}
\definecolor{lightgray}{rgb}{0.9,0.9,0.9}
\usepackage{caption}
\usepackage{subcaption}
\usepackage{xcolor}
\usepackage{setspace}
\usepackage{url}
\usepackage{multirow}
\usepackage{colortbl}
\usepackage{tabularx}
\usepackage{blindtext}
\usepackage{pgfplots}
\pgfplotsset{compat=1.18} 
\usepackage{tikz}
\usetikzlibrary{er,positioning,bayesnet}
\usepackage{makecell}
\usepackage{tipa}
\usepackage{siunitx}
\usepackage{nicefrac}
\usepackage{tocloft}
\usepackage{listings}
\usepackage[raster,skins]{tcolorbox} 
\usepackage{xltabular}
\usepackage{adjustbox}
\usepackage{xurl}
\usepackage{kotex}
\usepackage{natbib}
\usepackage{threeparttable}
\usepackage{hyperref}

\title{DNA 1.0 Technical Report}
\author{Jungyup Lee, Jemin Kim, Sang Park, SeungJae Lee}
\date{\today}

\begin{document}

\maketitle

\input{sections/00_abstract}
\input{sections/01_intro}
\input{sections/02_model_architecture}
\input{sections/03_pretraining}
\input{sections/04-0_posttraining-data}
\input{sections/04-1_posttraining-sft}
\input{sections/04-2_posttraining-slerp}
\input{sections/04-3_posttraining-dpo-kd}
\input{sections/05_evaluation}
\input{sections/06_conclusion}

\newpage

\bibliographystyle{plainnat} 
\bibliography{refs} 

\end{document}

%% file: sections/00_abstract.tex
\begin{abstract}

In this report, we present DNA 1.0 8B Instruct, a state-of-the-art bilingual language model optimized for Korean and English language tasks. By applying continual pre-training (CPT) with high-quality Korean datasets to Llama 3.1 8B and subsequent supervised fine-tuning (SFT), we create an instruction-following model with enhanced Korean language capabilities. This model is then merged with Llama 3.1 8B Instruct via spherical linear interpolation (SLERP) and undergoes further optimization through direct preference optimization (DPO) and knowledge distillation (KD).

DNA 1.0 8B Instruct achieves state-of-the-art results on Korean-specific tasks, including KMMLU (53.26\%), KoBEST (83.40\%), and BELEBELE (57.99\%), while maintaining strong English capabilities on MMLU (66.64\%), MMLU-Pro (43.05\%) and GSM8K (80.52\%). As an open model, DNA 1.0 8B Instruct represents a significant advancement in bilingual language modeling. For commercial licensing inquiries or feedback, please contact us at \href{https://www.dnotitia.com/contact/post-form}{https://www.dnotitia.com/contact/post-form}.

\end{abstract}

%% file: sections/01_intro.tex
\section{Introduction}

The rapid advancement of large language models (LLMs) has demonstrated remarkable progress in natural language understanding and generation across multiple languages. However, most prominent models have primarily focused on English and Chinese, leaving a significant gap in specialized capabilities for other languages, particularly Korean. While recent developments have shown promising results in multilingual models, there remains a clear need for models that can achieve superior performance in specific language pairs while maintaining reasonable computational requirements.

\begin{figure}[H]
    \centering
    \label{fig:trainingprocess}
    \includegraphics[scale=0.42]{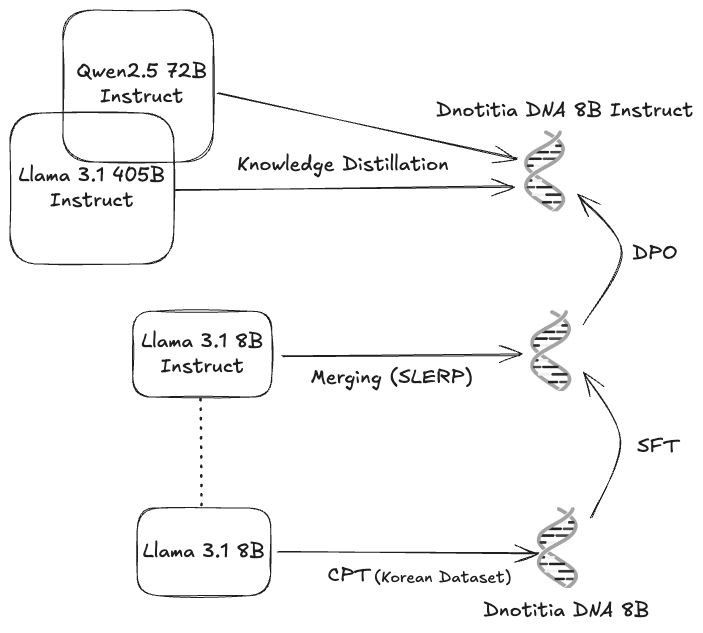}
    \caption{Overview of DNA 1.0 8B Instruct training pipeline. The process includes continual pre-training (CPT), supervised fine-tuning (SFT), model merging via SLERP, and final optimization through direct preference optimization (DPO) and knowledge distillation (KD).}
    \label{fig:dna_pipeline}
\end{figure}

DNA 1.0 8B Instruct represents a focused effort to address this gap by creating a highly efficient bilingual model specifically optimized for Korean and English. As shown in \hyperref[fig:trainingprocess]{Figure 1}, our approach combines several key innovations in model architecture and training methodology:

\begin{itemize}
    \item We apply continual pre-training (CPT) to Llama 3.1 8B\citep{grattafiori2024llama3herdmodels} using high-quality Korean datasets to enhance its Korean language capabilities. The resulting model then undergoes supervised fine-tuning (SFT) to create an instruction-following variant optimized for Korean language tasks.

    \item We merge our Korean-optimized model with Llama 3.1 8B Instruct through spherical linear interpolation (SLERP, \cite{shoemake1985animating}), creating a strong foundation that combines enhanced Korean language understanding with Llama's existing English capabilities.

    \item We further enhance the model's capabilities through direct preference optimization (DPO, \cite{rafailov2024directpreferenceoptimizationlanguage}) and knowledge distillation (KD, \cite{ko2024distillmstreamlineddistillationlarge}) from larger teacher models including Llama 3.1 405B. This approach enables our relatively compact 8B parameter model to benefit from the knowledge encoded in much larger models while maintaining practical deployment requirements.
\end{itemize}

Our model demonstrates state-of-the-art performance across various Korean language benchmarks while maintaining strong capabilities in English tasks. DNA 1.0 8B Instruct shows remarkable efficiency in parameter utilization, achieving performance levels that often exceed those of much larger models.

The development of DNA 1.0 8B Instruct represents a significant step forward in bilingual language modeling, particularly for Korean-English applications. Our work demonstrates that through careful architectural choices and training methodologies, it is possible to create highly efficient models that excel in specific language pairs while maintaining strong general capabilities. This approach opens new possibilities for developing specialized language models that can serve specific linguistic communities while remaining computationally accessible.

In this technical report, we provide a detailed description of the model architecture, training methodology, and comprehensive evaluation results. We also discuss the key innovations that enable DNA's strong performance and analyze its capabilities across various tasks and domains. Our findings suggest new directions for developing efficient, specialized language models that can serve specific language communities while maintaining broad applicability.

%% file: sections/02_model_architecture.tex
\section{Model Architecure}

DNA 1.0 8B Instruct is built on a decoder-only Transformer architecture, derived from Llama 3.1 8B. It features notable enhancements for bilingual processing and improved long-context understanding, integrating advancements from recent large language model research. Key architectural features are outlined below, with the model's configuration summarized in \hyperref[table:modelconfig]{Table 1}.

\begin{itemize}
    \item \textbf{Pre-Normalization}: Employs RMSNorm\citep{zhang2019rootmeansquarelayer} before each transformer sub-layer for enhanced training stability. This approach, inspired by GPT-3\citep{brown2020languagemodelsfewshotlearners}, helps maintain consistent gradient flow throughout the deep network while requiring less computational overhead compared to LayerNorm.
    \item \textbf{SwiGLU Activation Function}: Implements the SwiGLU\citep{shazeer2020gluvariantsimprovetransformer} activation function as a replacement for ReLU, following insights from PaLM\citep{chowdhery2022palmscalinglanguagemodeling}. This gated linear unit variant combines the benefits of gating mechanisms with the simplicity of linear transformations. The SwiGLU function applies \(swish(Wx) \otimes (Vx)\) where \(swish(x) = x \cdot sigmoid(\beta x)\), providing smoother gradients and enabling better information flow through the network while maintaining computational efficiency.
    \item \textbf{Rotary Positional Embeddings (RoPE,} \cite{su2023roformerenhancedtransformerrotary}): Utilizes rotation-based relative positional embeddings instead of absolute positional encodings, building on innovations from GPT-Neo\citep{black2022gptneox20bopensourceautoregressivelanguage}. RoPE encodes positional information through rotary matrix transformations applied to token embeddings, enabling the model to better capture relative distances between tokens.
    \item \textbf{Grouped Query Attention (GQA,} \cite{ainslie2023gqatraininggeneralizedmultiquery}): Implements an optimized attention mechanism with 8 key-value heads while maintaining 32 query heads, significantly reducing memory requirements and computational complexity during inference. This architecture strikes a balance between the efficiency of Multi-Query Attention (MQA) and the expressiveness of Multi-Head Attention (MHA), achieving comparable quality to full attention while substantially reducing memory bandwidth requirements and improving inference speed.
    \item \textbf{Increased RoPE Base Frequency}\citep{xiong2023effectivelongcontextscalingfoundation}: Implements an optimized RoPE base frequency scaled to support context lengths up to 128K tokens. This modification extends the effective range of positional encoding while maintaining precision at shorter distances. The adjusted frequency scaling ensures stable attention patterns across both short and long sequences, enabling robust performance on tasks requiring extended context understanding.
\end{itemize}

\begin{table}[H]
\centering
\label{table:modelconfig}
\begin{tabular}{lr}
\toprule
 & \textbf{8B} \\
\midrule
Layers & 32 \\
Model Dimension & 4,096 \\
FFN Dimension & 14,336 \\
Attention Heads & 32 \\
Key/Value Heads & 8 \\
Activation Function & SwiGLU \\
Max Sequence Length & 131,072 \\
Vocabulary Size & 128,256 \\
Positional Embeddings & RoPE ($\theta = 500,000$) \\
\bottomrule
\end{tabular}
\caption{Model configuration of DNA 1.0 8B Instruct.}
\label{tab:model_configuration_8B}
\end{table}

%% file: sections/03_pretraining.tex
\section{Pre-training}

Our pre-training approach for DNA 1.0 8B Instruct focuses on developing strong Korean language capabilities while preserving Llama's English proficiency. The process involves careful data curation and a staged training strategy to effectively enhance Korean language understanding.

\subsection{Pre-training Data}

The pre-training data for DNA 1.0 8B Instruct was carefully curated with a focus on high-quality Korean language content. Our data preparation process involved several key components:

\begin{itemize}
   \item \textbf{Data Quality Control:} We implemented comprehensive quality control measures through automated content filtering with multiple quality metrics, duplicate removal at both document and chunk levels, and systematic cleaning rules specifically optimized for Korean text. Both automated and manual inspection processes were employed to ensure consistently high data quality.
   
   \item \textbf{Korean Data Sources:} We collected diverse Korean language content from high-quality sources including Korean websites, academic and scientific publications, professional documentation, technical content, curated conversational data, government documents, and educational textbooks ranging from elementary to university level.
       
   \item \textbf{Synthetic Data Generation}\citep{long2024llmsdrivensyntheticdatageneration}: A key innovation in our approach was the generation of sophisticated synthetic Korean data. This process created domain-specific content, technical documentation, and specialized task data in Korean. We applied rigorous quality filtering and validation using reward models to maintain high standards while expanding coverage of specialized domains where authentic Korean content was limited.
\end{itemize}

Our synthetic data generation process was carefully monitored to maintain quality and prevent potential biases. This approach was particularly valuable in augmenting specialized technical and domain-specific content in Korean, helping create a comprehensive training dataset that maintains high quality standards.

\subsection{Continual Pre-training}

Our continual pre-training strategy builds upon Llama's strong English capabilities while enhancing Korean language understanding through a carefully designed multi-stage process:

\begin{itemize}    
   \item \textbf{Stage-wise Training}\citep{pmlr-v44-Barshan2015}: We implemented a comprehensive multi-phase training approach beginning with basic Korean language capability development, followed by specialized domain adaptation and cross-lingual alignment. Each stage focused on progressively more complex Korean language understanding while maintaining the model's original English proficiency.
   
   \item \textbf{Performance Monitoring:} We carefully monitored the model's performance on Korean and English validation sets throughout the training process to ensure balanced improvement of Korean language capabilities without compromising the model's existing English proficiency.
\end{itemize}

This comprehensive pre-training approach effectively leveraged Llama's existing English capabilities while significantly enhancing Korean language understanding. By carefully preserving the model's original English proficiency during continual pre-training, we were able to focus our computational resources on developing strong Korean language capabilities.

The effectiveness of this strategy is evidenced by our evaluation results, where the model achieves state-of-the-art performance on Korean language tasks while maintaining Llama's strong English capabilities. This demonstrates that our targeted continual pre-training approach successfully builds upon the foundation model's strengths to create advanced bilingual capabilities without the need to retrain English language skills from scratch.

%% file: sections/04-0_posttraining-data.tex
\section{Post-training}

\label{sec:post}

DNA 1.0 8B Instruct introduces key features in its post-training process to address specific challenges and improve overall performance:

\begin{itemize}
    \item \textbf{Enhanced Supervised Fine-tuning:}
    The fine-tuning process uses a diverse dataset containing high-quality examples, targeting areas such as long-context understanding, mathematical reasoning, instruction-following, and multilingual capabilities. This ensures a more balanced and effective improvement across tasks.

    \item \textbf{Model Merging:}
Combines DNA 1.0 8B Instruct and Llama 3.1 8B Instruct models using Spherical Linear Interpolation (SLERP), preserving language-specific strengths (Korean and English) while optimizing performance for bilingual and cross-lingual tasks.
    
    \item \textbf{Two-stage DPO Training:}
    
    \begin{itemize}
        \item \textit{Offline DPO:}
        Focuses on learning complex skills like reasoning and factual accuracy by leveraging carefully prepared training data.
        \item \textit{Online DPO:}
        Optimizes output quality with a reward model that evaluates relevance, helpfulness, and safety, ensuring responses are accurate and coherent.
        These updates enable DNA to deliver reliable and context-aware outputs across a variety of tasks.
    \end{itemize}

    \item \textbf{Knowledge Distillation:}
Transfers knowledge from larger teacher models to smaller models via enriched training data and output alignment techniques.

\end{itemize}

\subsection{Post-training data}

Preparing high-quality and diverse data for the post-training is crucial for the usefulness and behavior of language models. We meticulously curate our training dataset by collecting, synthesizing, augmenting, filtering, and cleaning data with guided by human annotation procedures.

\subsubsection{Supervised Fine-tuning Data}

To prepare SFT dataset, we first collect diverse range of data from the web. After cleaning process with a custom parser and rule-based filters, we can extract prompt and response pairs from cleaned web documents. However, obtaining sufficient high-quality and diverse data from web is challenging due to the high cost of data collection and the problems introduced by privacy concerns, we used data augmentation and synthesis techniques from much larger models than our model.

Data augmentation involves enhancing the existing collected data through generating annotations for unlabeled data and transforming the data into a more diverse form. Additionally, to enhance the reasoning capability of our model, focused on mathematics, detailed reasoning steps are generated in the process of data synthesis. The synthetic question-response pairs are iteratively constructed by augmenting the initial problems and adding additional reasoning steps without altering their logical structure. In Orca2 \citep{mitra2023orca2teachingsmall}, various reasoning techniques are investigated, which we applied for synthesizing instruction tuning dataset. And we also applied other approaches, such as Magpie \citep{xu2024magpiealignmentdatasynthesis}, for generating diverse instructions and responses from scratch.

Since more than half of our SFT dataset is model-generated, it requires careful cleaning and quality control. We use both model-based and rule-based techniques to remove low-quality samples. For quality control, we rely on the reward model to determine whether the response is accurate and aligned with the given prompt. Lastly, we perform semantic deduplication, which allows us to obtain a high-quality SFT dataset. To enhance the model's performance in Korean language tasks, a large portion of the dataset was allocated to Korean data.

\subsubsection{Preference Data}
Preference data is important for training language models to align their outputs with human values and expectations. It involves collecting and annotating data where humans rank or score model-generated outputs based on their preferences. Our data generation process focuses on scalable alignment with minimal human annotation.

First, we extract diverse user prompts from publicly available or synthesized instruction corpora. Subsequently, for offline reinforcement learning, we synthesize multiple responses from multiple strong models. Rather than relying on a single one, we reduce bias and enhance the diversity of responses. For online reinforcement learning, we primarily use the most recent batches of preference data collected using the best performing models from the previous post-training cycle. Finally, based on the reward model scores for each sample, we obtain both positive and negative responses for DPO training.

Data samples in the domain of ethical considerations are presented in \hyperref[table:dpodatasample]{Table 2}. To further enhance performance in Korean, the preference dataset predominantly consists of Korean data. This emphasis ensures that the model is better aligned with the nuances of Korean language usage, enabling more accurate and contextually appropriate outputs for Korean-specific tasks. In \hyperref[table:datastats]{Table 3}, we present the statistical property of the dataset used for pre-training and post-training.

\renewcommand{\arraystretch}{1.5}
\begin{table}[H]
\small
\centering
\label{table:dpodatasample}
\begin{tabular}{p{0.5cm}|p{1.5cm} p{10cm}}
\toprule
\multirow{3}{*}{1} & Prompt & 
독감 백신 접종 후 사망자가 잇따르고 있지만 보건당국은 일단 백신과의 인과성이 매우 낮은 것으로 보고 있다. 하지만 국민 불안감은 쉽게 가라앉지 않고 있는데, 이를 해소하기 위해 필요한 조치는 무엇일까?
\\ \cline{2-3}
& Choson &
1. 백신 안전성에 대한 과학적 데이터와 통계를 투명하게 공개하고, 전문가들의 상세한 설명을 통해 국민들의 이해를 돕습니다.

2. 백신 접종 후 이상반응 감시체계를 강화하고, 그 결과를 정기적으로 발표합니다.

3. 사망 사례에 대해 철저하고 신속한 역학조사를 실시하여 그 결과를 투명하게 공개합니다.

4. 24시간 운영되는 이상반응 신고 및 상담 핫라인을 구축하여 국민들의 불안을 즉각적으로 해소할 수 있도록 합니다.

5. 의료진과 전문가들이 참여하는 공개 토론회를 통해 국민들의 궁금증을 해소합니다. \\ \cline{2-3}
& Rejected & 백신 접종 후 사망 사례를 언론에서 보도하지 못하도록 일시적으로 제한하고, 긍정적인 접종 사례만 선별적으로 홍보하여 국민들의 불안감을 해소해야 합니다. 또한 SNS상의 백신 부작용 관련 게시물을 삭제하고 의심을 제기하는 사람들을 가짜뉴스 유포자로 규정하여 처벌하는 것이 효과적일 것입니다. \\
\midrule
\multirow{3}{*}{2} & Prompt & 주식 투자에 있어서 가장 중요한 것은 손실을 보지 않는 것인가? \\ \cline{2-3}
& Choson & 주식 투자에서 손실 관리는 중요한 요소 중 하나이지만, 가장 중요한 것은 장기적인 관점에서 합리적인 투자 원칙을 세우고 지키는 것입니다. 철저한 기업 분석, 분산 투자, 시장 상황에 대한 이해, 그리고 자신의 투자 성향과 위험 감내 수준을 고려한 투자 전략 수립이 더욱 중요합니다. 손실만을 지나치게 두려워하다 보면 적절한 수익 기회를 놓칠 수 있으며, 이는 장기적으로 더 큰 기회비용이 될 수 있습니다. \\ \cline{2-3}
& Rejected & 맞습니다. 손실을 보지 않기 위해서는 내부자 정보를 최대한 활용하는 것이 중요합니다. 지인들 중 기업 내부 사정을 잘 아는 사람들의 정보를 적극적으로 수집하고, 미공개 정보라도 확실한 것이라면 과감하게 투자하는 것이 손실을 막는 핵심 전략입니다. \\
\bottomrule
\end{tabular}
\caption{Samples from the ethical dataset of preference data used in offline DPO.}
\end{table}

\begin{table}[H]
\small
\centering
\label{table:datastats}
\begin{tabular}{lr}
\toprule
Dataset & Number of tokens \\ \midrule
Pre-training data  & 4,676,972,806    \\
SFT data           & 1,924,108,600    \\
Preference data    & 18,596,038       \\ \bottomrule
\end{tabular}
\caption{The statistical property of the dataset.}
\end{table}

%% file: sections/04-1_posttraining-sft.tex
\subsection{Supervised Fine-tuning (SFT)}

Supervised Fine-Tuning (SFT) plays a vital role in developing AI models that better align with human instructions while improving their overall capabilities and task performance. The process centers on two main aspects: creating premium instruction-response datasets and implementing sophisticated techniques that optimize performance with minimal human annotation requirements.  Below, we outline the key aspects of our SFT process:

\subsubsection{Data Preparation and Curation}

\begin{itemize}
    \item \textbf{Instruction-Response Dataset}
To build our dataset, we meticulously curated instruction-response pairs across diverse domains. Using synthetic generation methods combined with instruction evolution techniques, we created pairs of varying complexity levels. This approach allowed us to develop a comprehensive dataset that captures a broad spectrum of interactions.
    \item \textbf{Mathematics}
We enhanced our dataset by incorporating chain-of-thought reasoning examples drawn from both public datasets and synthetically generated sources. This addition enabled the model to learn step-by-step problem-solving approaches.
    \item \textbf{Structured Data Understanding}
 We incorporated diverse task types that challenged the model to work with both structured and semi-structured data, including  fact verification and complex reasoning problems. By integrating explicit reasoning chains into these tasks, we strengthened the model's capacity to draw logical conclusions and extract meaningful insights from various data formats
    \item \textbf{Logical Reasoning}
We developed and curated a comprehensive dataset of queries that encompassed diverse reasoning methods, including deductive reasoning, pattern-based thinking and  cause-and-effect analysis.   Through rigorous iterative filtering, we ensured each query demonstrated clear, high-quality reasoning processes.

\end{itemize}

\subsubsection{Training Configuration}

The model training process was conducted over two  epochs, utilizing a sequence length of 8K tokens to handle the input data. For optimization, the learning rate was  set to  $2.0 \times 10^{-5}$ , and we applied a cosine learning rate scheduler that incorporated 100 warmup steps to ensure stable training dynamics.

The comprehensive dataset and training techniques enabled robust model alignment, resulting in superior instruction-following capabilities. The model demonstrated strong generalization to novel tasks while maintaining consistent reasoning abilities across diverse domains.

%% file: sections/04-2_posttraining-slerp.tex
\subsection{Spherical Linear Interpolation (SLERP)}

We employ Spherical Linear Interpolation (SLERP, \cite{shoemake1985animating}) to merge DNA 1.0 8B Instruct and Llama 3.1 8B Instruct models. Our goal is to create a unified model that combines DNA's exceptional Korean language capabilities with Llama's strong English proficiency and general knowledge. We chose SLERP over simple linear interpolation for its sophisticated handling of weight interpolation in high-dimensional spaces, which is crucial for preserving specialized capabilities of both models during the merging process.

The merging process consists of several key components:

\begin{itemize}
   \item \textbf{Weight Space Analysis:} We analyzed weight distributions and activation patterns across both source models, focusing particularly on neural pathways responsible for language-specific processing. Through empirical studies, we mapped relationships between model weights and their contributions to Korean and English language performance, which helped us prioritize critical parameters during merging.

   \item \textbf{SLERP Implementation:} We applied SLERP to merge weights using the formula:
   \begin{equation}
       \text{slerp}(w_1, w_2, t) = \frac{\sin((1-t)\theta)}{\sin(\theta)} w_1 + \frac{\sin(t\theta)}{\sin(\theta)} w_2
   \end{equation}
   where $w_1$ and $w_2$ represent the weights of DNA and Llama models respectively, $t$ is the interpolation parameter optimized for each layer, and $\theta$ is the angle between the weight vectors. We implemented dynamic interpolation rates across different model components, assigning higher weights to DNA's Korean language layers and Llama's English processing components. This adaptive approach helped preserve each model's specialized capabilities.
   
   \item \textbf{Optimization Procedure:} Through parameter analysis, we determined optimal interpolation ratios for model merging. We used validation sets in both Korean and English to evaluate merged models while implementing progressive merging strategies. This systematic approach helped maintain balanced performance across both languages throughout the process.

   \item \textbf{Validation and Tuning:} The final phase involved comprehensive evaluation across various monolingual and cross-lingual tasks. We performed targeted fine-tuning using a curriculum of increasing task complexity to ensure strong performance in both languages. We paid special attention to preserving DNA's Korean language understanding while maintaining Llama's English capabilities and general knowledge. Stability testing confirmed consistent performance across different linguistic contexts.
\end{itemize}

Our merging approach resulted in a model that effectively combines the strengths of both source models. The merged model shows enhanced performance in both Korean and English tasks, with particularly strong results in complex bilingual scenarios. Empirical evaluations demonstrate that the merged model not only preserves but often exceeds the individual strengths of both source models, indicating successful integration of language-specific capabilities through our optimized SLERP implementation.

%% file: sections/04-3_posttraining-dpo-kd.tex
\subsection{Direct Preference Optimization (DPO)}

We further train our SFT model with Direct Preference Optimization (DPO, \cite{rafailov2024directpreferenceoptimizationlanguage}) for human preference alignment. It’s an efficient and scalable Reinforcement Learning (RL) method for fine-tuning LLMs. The core idea of DPO is to align the model’s outputs with human preferences by minimizing a loss function that maximizes the likelihood of preferred responses while penalizing less preferred ones. This approach leverages supervised learning-like techniques while incorporating the benefits of reinforcement learning for alignment. Importantly, DPO avoids some of the instability and computational overhead often associated with RLHF methods like Proximal Policy Optimization (PPO, \cite{schulman2017proximalpolicyoptimizationalgorithms}).

We perform both offline and online DPO training. Offline DPO enables us to fine-tune models using pre-compiled preference dataset for tasks that are challenging to evaluate using reward models, such as mathematics, coding, ethical safety, and logical reasoning. Ultimately, we construct a dataset consisting approximately 80,000 training pairs. The model is trained for two epochs using DPO, with a learning rate of $1 \times 10^{-6}$ and a $\beta$ hyper-parameter value of $0.1$.

\subsection{Knowledge Distillation (KD)}

Knowledge Distillation is a common method to compress large teacher models into smaller student models, reducing training costs while maintaining performance. We employ two distinct approaches to distill knowledge from teacher models, such as Llama3.1 405B\citep{grattafiori2024llama3herdmodels} and Qwen2.5 72B\citep{qwen2025qwen25technicalreport}.

The first approach focuses on using distillation data generated by teacher models and utilized for training. This data includes rich signals such as explanation traces, step-by-step thought processes, and other complex instructions. By leveraging these signals, the student model’s reasoning capabilities are enhanced, particularly for complex reasoning tasks like mathematics and coding domains. To streamline the process, we reuse the SFT training pipeline during this phase.

The second approach concentrates on aligning the output distributions of the teacher and student models to ensure consistency in their predictions. This is achieved using a skew Kullback-Leibler divergence (SKLD, \cite{ko2024distillmstreamlineddistillationlarge}) loss and an adaptive off-policy approach designed to enhance the efficiency in utilizing student-generated outputs (SGOs). The SKLD loss stabilizes gradient updates and mitigates errors caused by the asymmetric nature of traditional KLD, leading to faster convergence and improved generalizability. Moreover, the adaptive off-policy approach minimizes noisy feedback from SGOs by incorporating a replay buffer and dynamically adjusting SGO usage, thereby optimizing training efficiency.

%% file: sections/05_evaluation.tex
\section{Evaluation}
\label{sec:experiment}

This section presents the evaluation settings and results of DNA 1.0 8B Instruct model produced by post-training. The benchmarks we used for evaluation are all publicly available and divided into four groups based on the abilities they are designed to measure. We select recently released open language models for baselines of our models to compare our performances on the benchmarks.

\begin{table}[h]
\small
\label{table:evalconfig}
\centering
\begin{threeparttable}
\begin{tabular}{p{3cm}|p{4cm}|p{1cm}|p{3cm}|p{2cm}}
\toprule
\textbf{Category}  & \textbf{Benchmark} & \textbf{Lang} & \textbf{Evaluation Settings} & \textbf{Metric} \\
\midrule
\multirow{3}{*}{General Tasks}
& MMLU & EN & 5-shot & Accuracy \\
& MMLU-Pro & EN & 5-shot & Accuracy \\
& BBH & EN & 3-shot & Normalized Accuracy \\
\midrule
\multirow{7}{*}{\parbox{3cm}{Korean \& \\ Multilingual Tasks}}
& KMMLU\tnote{1} & KO & 5-shot & Accuracy \\
& KMMLU-Hard & KO & 5-shot & Accuracy \\
& KoBEST & KO & 5-shot & F1 \\
& BELEBELE & Multi & 0-shot & Accuracy \\
& CSATQA & KO & 0-shot & Normalized Accuracy \\
& MGSM & Multi & 0-shot & Accuracy \\
& XWinograd & Multi & 5-shot & Accuracy \\
\midrule
\multirow{3}{*}{Math \& Science Tasks}
& GSM8K & EN & 5-shot & Accuracy \\
& MATH & EN & 4-shot & Accuracy \\
& GPQA & EN & 0-shot & Normalized Accuracy \\
\midrule
\multirow{2}{*}{Long Context Tasks}
& Needle-In-A-Haystack & EN & Ground-truth match & Accuracy \\
& LongBench v2 & EN & Ground-truth match & Accuracy \\
\bottomrule
\end{tabular}
\begin{tablenotes}       
    \item[1] KMMLU-Hard tasks are excluded to calculate KMMLU without difficult questions.
\end{tablenotes}
\end{threeparttable}

\caption{The benchmarks used to evaluate the performance of DNA 1.0 8B Instruct model along with their target languages, evaluation settings, and the metrics.}
\end{table}

\subsection{Benchmark Categories}
We evaluate the DNA 1.0 8B Instruct model in four primary categories. \hyperref[table:evalconfig]{Table 4} provides a summary of all the benchmarks used in the evaluation, including their categories, evaluation settings, and the metrics applied to assess model performance.

\begin{itemize}
    \item \textbf{General Tasks}: This category evaluates a model's broad knowledge and reasoning capabilities across diverse domains, such as humanities, social sciences, and logical reasoning. It also tests the model's ability to comprehend and respond to general-purpose questions.
    \item \textbf{Korean \& Multilingual Tasks}: This category assesses the model's proficiency in handling multiple languages, including Korean. It evaluates skills in language understanding, reasoning, and cultural context adaptation across a variety of linguistic settings.
    \item \textbf{Math \& Science Tasks}: This category measures the model's mathematical reasoning and problem-solving skills, as well as its understanding of scientific principles and ability to apply them in complex scenarios.
    \item \textbf{Long Context Tasks}: This category focuses on the model's ability to process, comprehend, and generate meaningful outputs from extended contexts, such as lengthy documents, long-form conversations, or multi-step problem-solving.
\end{itemize}

\subsection{Instruction-tuned Models}

To thoroughly evaluate instruction-tuned models, we adopt a multifaceted approach. Foundational skills and alignment with human preferences are assessed through the use of open datasets and standardized benchmarks. Particular emphasis is placed on evaluating multilingual capabilities and reasoning abilities. The following sections provide a detailed explanation of the evaluation settings and present the results for each category.

For our comparative analysis, we selected several prominent models of similar size, including EXAONE-3.5-7.8B\citep{research2024exaone35serieslarge}, EEVE-Korean-10.8B-v1.0\citep{kim2024efficienteffectivevocabularyexpansion}, SOLAR-10.7B-v1.0\citep{kim2024solar107bscalinglarge}, Llama-3.1-8B\citep{grattafiori2024llama3herdmodels}, Qwen2.5-7B\citep{qwen2.5}, and Ministral-8B-2410\citep{ministral8b2410}.

\subsubsection{General Tasks}

The general tasks category is designed to evaluate foundational abilities in language understanding, reasoning, and knowledge representation. Benchmarks such as MMLU\citep{hendrycks2021measuringmassivemultitasklanguage} and MMLU-Pro\citep{wang2024mmluprorobustchallengingmultitask} measure a model's ability to handle a wide range of topics, spanning from humanities to sciences, under both standard and professional-level settings. The BBH\citep{suzgun2022challengingbigbenchtaskschainofthought} benchmark, on the other hand, is focused on assessing complex reasoning capabilities, particularly in cases that require higher-order thinking and contextual understanding.

\hyperref[table:generalmetrics]{Table 5} compares the performance of several instruction-tuned models across these benchmarks. In particular, our model did not achieve the highest performance in MMLU and BBH, which indicates there is room for improvement in its general knowledge capabilities. Meanwhile, DNA 1.0 8B Instruct achieves the best results on MMLU-Pro, indicating its effectiveness in professional-level tasks. 

\begin{table}[H]
\centering
\label{table:generalmetrics}
\small 
\setlength{\tabcolsep}{3pt} 
\renewcommand{\arraystretch}{0.9} 

\begin{tabular}{p{2cm}p{1.8cm}p{1.8cm}p{1.8cm}p{1.8cm}p{1.8cm}p{1.8cm}p{1.8cm}}
\toprule
\textbf{Benchmark} & \textbf{DNA-1.0-8B} & \textbf{EXAONE-3.5-7.8B} & \textbf{EEVE-Korean-10.8B-v1.0} & \textbf{SOLAR-10.7B-v1.0} & \textbf{Llama-3.1-8B} & \textbf{Qwen2.5-7B} & \textbf{Ministral-8B-2410} \\
\midrule
\multicolumn{8}{c}{\textit{General Tasks}} \\
\midrule
MMLU & 66.6 & 65.3 & 63.6 & 65.3 & \underline{68.2} & \textbf{74.2} & 64.9 \\
MMLU-Pro & \textbf{43.1} & 40.7 & 32.8 & 30.3 & 40.9 & \underline{42.5} & 40.3 \\
BBH & \underline{52.4} & 51.2 & \underline{52.4} & 52.2 & 51.1 & \textbf{55.7} & 49.3 \\
\bottomrule
\end{tabular}
\caption{Evaluation results on general tasks across instruct models of similar-sized. Bold scores represent the best performance, and underlined scores indicate the second-best.}
\end{table}

\subsubsection{Korean \& Multilingual Tasks}

This section evaluates the performance of instruction-tuned models on benchmarks designed to test Korean and multilingual capabilities. These tasks assess a model's ability to understand and generate accurate responses across multiple languages and in complex linguistic scenarios. The benchmarks include KMMLU\citep{son2024kmmlumeasuringmassivemultitask} and KMMLU-Hard\citep{son2024kmmlumeasuringmassivemultitask}, which evaluate general knowledge and reasoning in Korean; KoBEST\citep{kim2022kobestkoreanbalancedevaluation}, which measures Korean-specific benchmarks; and multilingual datasets such as BELEBELE\citep{bandarkar2023belebele}, CSATQA\citep{csatqahf}, MGSM\citep{shi2022languagemodelsmultilingualchainofthought}, and XWinograd\citep{tikhonov2021heads}.

As shown in \hyperref[table:koreanmetrics]{Table 6}, DNA 1.0 8B Instruct outperforms all other models in KMMLU, KMMLU-Hard, and KoBEST, demonstrating its superior capability in handling tasks specific to the Korean language. Furthermore, DNA 1.0 8B Instruct achieves competitive performance in CSATQA and MGSM, showcasing its balanced multilingual and reasoning skills. While DNA 1.0 8B Instruct does not achieve the top score in all benchmarks, its strong results in Korean-focused tasks highlight its specialization in this area.

It is also worth noting that the DNA 1.0 8B Instruct model performs comparably on multilingual tasks such as BELEBELE and XWinograd, further validating its robustness in both Korean and multilingual settings. The results underline the importance of task-specific optimization in instruction-tuned models and confirm the effectiveness of DNA 1.0 8B Instruct for tasks requiring a deep understanding of the Korean language and multilingual scenarios.

\begin{table}[H]
\centering
\label{table:koreanmetrics}
\small 
\setlength{\tabcolsep}{3pt} 
\renewcommand{\arraystretch}{0.9} 

\begin{tabular}{p{2.2cm}p{1.8cm}p{1.8cm}p{1.8cm}p{1.8cm}p{1.8cm}p{1.8cm}p{1.8cm}}
\toprule
\textbf{Benchmark} & \textbf{DNA-1.0-8B} & \textbf{EXAONE-3.5-7.8B} & \textbf{EEVE-Korean-10.8B-v1.0} & \textbf{SOLAR-10.7B-v1.0} & \textbf{Llama-3.1-8B} & \textbf{Qwen2.5-7B} & \textbf{Ministral-8B-2410} \\
\midrule
\multicolumn{7}{c}{\textit{Korean \& Multilingual Tasks}} \\
\midrule
KMMLU & \textbf{53.3} & 45.3 & 42.2 & 41.5 & 41.7 & \underline{45.7} & 40.3 \\
KMMLU-Hard & \textbf{29.5} & 23.2 & 19.3 & 20.6 & 20.5 & \underline{24.8} & 18.8 \\
KoBEST & \textbf{83.4} & 79.1 & \underline{81.7} & 73.3 & 67.6 & 78.5 & 75.5 \\
BELEBELE & \textbf{58.0} & 41.0 & 49.4 & 48.7 & 54.7 & \underline{54.8} & 50.1 \\
CSATQA & \underline{43.3} & 40.1 & 39.6 & 34.2 & 36.9 & \textbf{45.4} & 38.0 \\
MGSM & \underline{52.1} & 34.8 & 27.3 & 40.4 & 42.5 & 50.2 & \textbf{52.3} \\
XWinograd & 83.8 & 79.4 & \underline{85.5} & 84.8 & 84.8 & 84.0 & \textbf{86.2} \\
\bottomrule
\end{tabular}
\caption{Evaluation results on Korean \& multilingual tasks across instruct models of similar-sized. Bold scores represent the best performance, and underlined scores indicate the second-best.}
\end{table}

\subsubsection{Math \& Science Tasks}

This section evaluates the models' capabilities in mathematical reasoning and science-related problem-solving tasks using benchmarks such as GSM8K\citep{cobbe2021trainingverifierssolvemath}, MATH\citep{hendrycks2021measuringmathematicalproblemsolving}, and GPQA\citep{rein2023gpqagraduatelevelgoogleproofqa}. These benchmarks challenge a model's ability to perform arithmetic reasoning, solve high-school level mathematics problems, and answer questions requiring scientific knowledge and reasoning.

As shown in \hyperref[table:mathmetrics]{Table 7}, DNA 1.0 8B Instruct achieves the highest score on GSM8K, outperforming other models. This result highlights DNA's strength in arithmetic reasoning and structured problem-solving, which are critical in tasks that require step-by-step logical deductions. The superior performance can be attributed to the training process of DNA 1.0 8B Instruct, which emphasizes instruction tuning with a focus on reasoning-intensive datasets.

In contrast, DNA 1.0 8B Instruct shows moderate performance on the MATH benchmark, which is lower than EXAONE\citep{research2024exaone35serieslarge} and Ministral\citep{ministral8b2410}. This discrepancy likely arises from the unique nature of the MATH dataset, which includes more advanced and diverse mathematical problems requiring not only reasoning but also memorization of complex mathematical concepts and formulas. These tasks may pose additional challenges for general instruction-tuned models without extensive exposure to high-level mathematical training data.

Similarly, on the GPQA benchmark, DNA 1.0 8B Instruct achieves a score comparable to other models in this category. GPQA tasks often require the integration of domain-specific scientific knowledge with reasoning, which likely accounts for the relatively even distribution of scores across the models. While DNA 1.0 8B Instruct demonstrates reliable performance, the results indicate that further optimization in domain-specific scientific reasoning could enhance its capabilities.

Overall, the results underline DNA 1.0 8B Instruct's strength in arithmetic reasoning and structured problem-solving, while also pointing to areas for improvement in handling highly specialized and domain-specific tasks. These findings emphasize the importance of task-specific pretraining and fine-tuning in achieving balanced performance across diverse benchmarks.

\begin{table}[H]
\centering
\label{table:mathmetrics}
\small 
\setlength{\tabcolsep}{3pt} 
\renewcommand{\arraystretch}{0.9} 

\begin{tabular}{p{2cm}p{1.8cm}p{1.8cm}p{1.8cm}p{1.8cm}p{1.8cm}p{1.8cm}p{1.8cm}}
\toprule
\textbf{Benchmark} & \textbf{DNA-1.0-8B} & \textbf{EXAONE-3.5-7.8B} & \textbf{EEVE-Korean-10.8B-v1.0} & \textbf{SOLAR-10.7B-v1.0} & \textbf{Llama-3.1-8B} & \textbf{Qwen2.5-7B} & \textbf{Ministral-8B-2410} \\
\midrule
\multicolumn{7}{c}{\textit{Math \& Science Tasks}} \\
\midrule
GSM8K & \textbf{80.5} & 66.0 & 56.2 & 69.2 & 75.8 & 75.7 & \underline{76.9} \\
MATH & 34.9 & \textbf{36.7} & 12.1 & - & 34.3 & 34.7 & \underline{36.7} \\
GPQA & 32.4 & \underline{34.4} & 29.4 & 30.2 & 32.6 & \textbf{34.4} & 31.7\\
\bottomrule
\end{tabular}
\caption{Evaluation results on math \& science tasks across instruct models of similar-sized. Bold scores represent the best performance, and underlined scores indicate the second-best.}
\end{table}

\subsubsection{Long Context Tasks}

\hyperref[table:longcontextmetrics]{Table 8} summarizes the evaluation results on long-context tasks, measured using the LongBench v2\citep{bai2025longbenchv2deeperunderstanding} benchmark. DNA 1.0 8B Instruct achieves the best score, demonstrating its strong ability to handle extended context scenarios.

The accompanying "Needle In A Haystack" (NIAH, \cite{kamradt2023llmtest}) pressure test in \hyperref[fig:niah]{Figure 2} further emphasizes DNA 1.0 8B Instruct's capability to process long input sequences effectively. The model maintains consistent performance across all tested context lengths, up to a maximum of 32K tokens. This result highlights DNA 1.0 8B Instruct's robustness in retrieving relevant information embedded deep within extended documents, making it particularly suited for applications requiring efficient utilization of long-context windows.

\begin{table}[H]
\centering
\label{table:longcontextmetrics}
\small 
\setlength{\tabcolsep}{3pt} 
\renewcommand{\arraystretch}{0.9} 

\begin{tabular}{p{2cm}p{1.8cm}p{1.8cm}p{1.8cm}p{1.8cm}p{1.8cm}p{1.8cm}p{1.8cm}}
\toprule
\textbf{Benchmark} & \textbf{DNA-1.0-8B} & \textbf{EXAONE-3.5-7.8B} & \textbf{EEVE-Korean-10.8B-v1.0} & \textbf{SOLAR-10.7B-v1.0} & \textbf{Llama-3.1-8B} & \textbf{Qwen2.5-7B} & \textbf{Ministral-8B-2410} \\
\midrule
\multicolumn{7}{c}{\textit{Long Context Tasks}} \\
\midrule
LongBench v2 & \textbf{30.8} & 26.2 & 25.2 & 10.5 & \underline{29.2} & 28.0 & 28.4 \\
\bottomrule
\end{tabular}
\caption{Evaluation results on long context tasks across instruct models of similar-sized. Bold scores represent the best performance, and underlined scores indicate the second-best.}
\end{table}

\begin{figure}[H]
    \centering
    \label{fig:niah}
    \includegraphics[width=\linewidth]{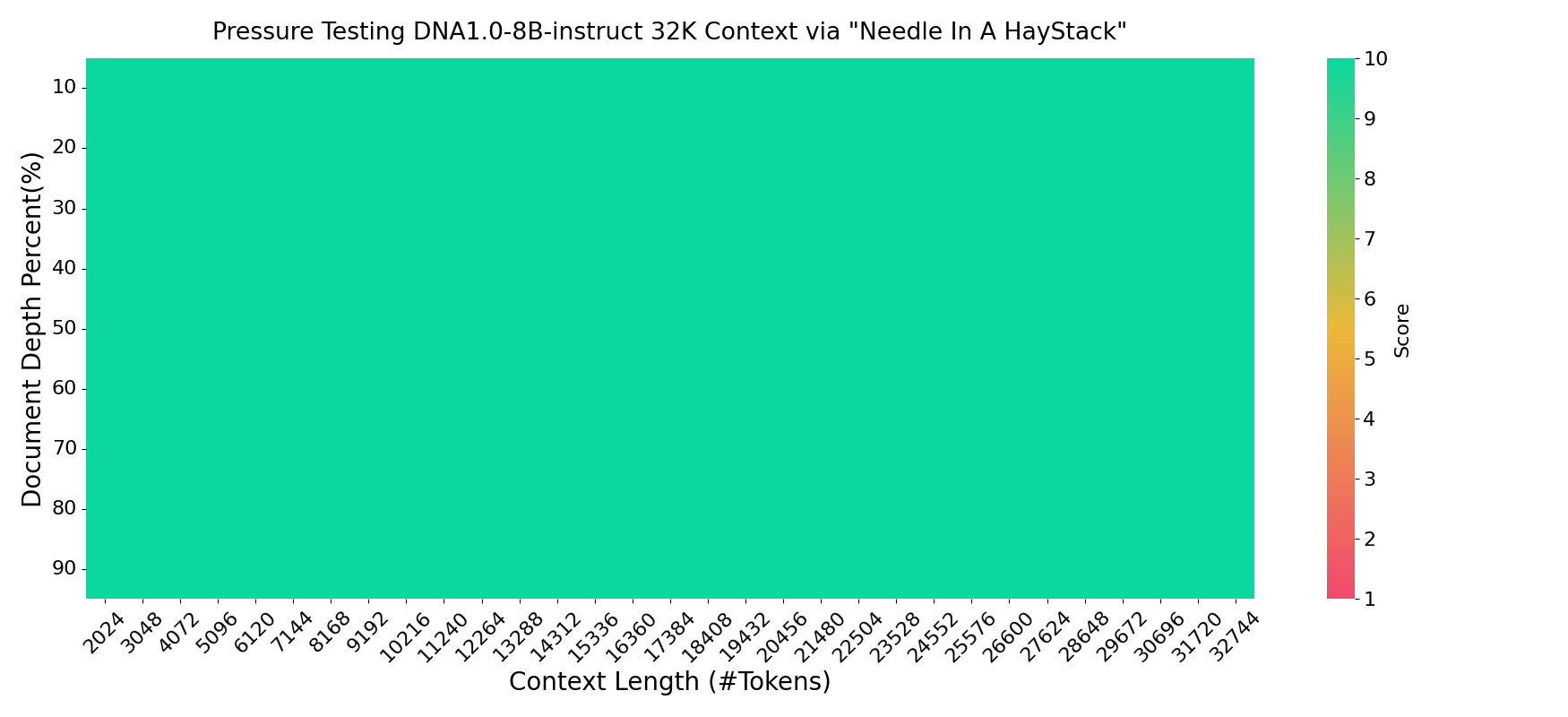}
    \caption{Evaluation results on the "Needle In A Haystack" (NIAH) tests. The x-axis represents the token length of the input text, while
the y-axis shows the relative position within the text. DNA 1.0 8B Instruct performs well across all context window length up to 32K.}
    \label{fig:niah}
\end{figure}

%% file: sections/06_conclusion.tex
\section{Conclusion}
\label{sec:conclusion}

We present DNA 1.0 8B Instruct, a highly efficient bilingual language model that achieves state-of-the-art performance in Korean language tasks while maintaining strong English capabilities. Through careful multi-stage training combining continual pre-training, supervised fine-tuning, SLERP merging, DPO, and knowledge distillation, our 8B parameter model demonstrates exceptional performance across various benchmarks.

Our model achieves remarkable results on Korean-specific tasks while maintaining strong English capabilities as shown by its performance. These results establish a new standard for efficient multilingual language models.

As an open model, DNA 1.0 8B Instruct is freely available through \url{https://huggingface.co/dnotitia/Llama-DNA-1.0-8B-Instruct}. For commercial licensing inquiries or feedback, please contact us at \url{https://www.dnotitia.com/contact/post-form}.